\documentclass[preprint,12pt,authoryear]{elsarticle}
\usepackage[margin=1in]{geometry}
\usepackage{url}            
\usepackage{booktabs}       
\usepackage{amsfonts}       
\usepackage{microtype}      
\usepackage{amsmath}
\usepackage{amssymb}
\usepackage{amsfonts}
\usepackage{bm}
\usepackage{url}
\usepackage{caption}
\usepackage{makecell}
\usepackage{multirow}
\usepackage{subcaption}
\usepackage{graphicx}
\usepackage[linesnumbered,ruled,vlined]{algorithm2e}

\newcommand{\splitcell}[1]{%
  \begin{tabular}{@{}c@{}}\strut#1\strut\end{tabular}%
}

\usepackage{lineno}
\journal{International Journal of Transportation Science and Technology}

\begin{document}
\begin{frontmatter}

\title{Graph Neural Networks for Travel Distance Estimation and Route Recommendation Under Probabilistic Hazards}

\author[inst1]{Tong Liu\footnote{Corresponding Author, tongl5@illinois.edu}}
\author[inst1]{Hadi Meidani}
\affiliation[inst1]{organization={University of Illinois, Urbana-Champaign, Department of Civil and Environmental Engineering},
            addressline={205 N Mathews Ave}, 
            city={Urbana},
            postcode={61801}, 
            state={IL},
            country={USA}}

\begin{abstract}
Estimating the shortest travel time and providing route recommendation between different locations in a city or region can quantitatively measure the conditions of the transportation network during or after extreme events. One common approach is to use Dijkstra's Algorithm, which produces the shortest path as well as the shortest distance. However, this option is computationally expensive when applied to large-scale networks. This paper proposes a novel fast framework based on graph neural networks (GNNs) which approximate the single-source shortest distance between pairs of locations, and predict the single-source shortest path subsequently. We conduct multiple experiments on synthetic graphs of different size to demonstrate the feasibility and computational efficiency of the proposed model. In real-world case studies, we also applied the proposed method of flood risk analysis of coastal urban areas to calculate delays in evacuation to public shelters during hurricanes. The results indicate the accuracy and computational efficiency of the GNN model, and its potential for effective implementation in emergency planning and management.
\end{abstract}


\end{frontmatter}

\section{Introduction}
\label{sec:introduction}
Extreme events like natural hazards have caused significant disruptions to infrastructure systems in urban areas for decades \citep{kuo2023public,liu2023physics}. For instance, Hurricane Irma severely impacted life and supply chains along the coastal area, specifically in Florida \citep{issa2018deaths}, resulting in a considerable amount of time for recovery from the disruption. The estimate for economic loss due to fatality, social disruption, and post-disaster reconstruction exceeded hundreds of billion dollars \citep{zhu2020estimating}. The transportation system, as one of the sixteen critical infrastructure systems \citep{cisa}, plays a significant role in facilitating recovery after natural hazards. The national highway system in the US, spanning over 164,000 miles, is relied upon during emergencies to maintain access to critical facilities such as police stations, public shelters, airports, hospitals, and fire departments \citep{liu2024neural}.

In order to assess the reliability roadway systems, one needs to evaluate the performance of roadway systems before, during and after natural disasters. This helps identify infrastructure bottlenecks, develop emergency plans, and effectively allocate resources. The roadway system reliability can be investigated considering different performance metrics. These include traffic flow distribution \citep{liu2024end}, accessibility \citep{liu2024graph}, and resilience index \citep{singh2021lessons}. Among the various performance metrics, the travel distance and travel time between locations in a roadway network can be an effective measure for emergency planning \citep{xu2023non, sun2024bayesian}. Specifically, the estimated distances in a disaster can help locate public shelters, estimate rescue time, and ensure public safety \citep{liao2022agent,fan2023climate}. Although other metrics, including mobility and accessibility, can provide more detailed information, they require additional information, such as origin-destination demand and traffic capacity, which may not be easily available due to sudden changes in transportation patterns after a disaster. 

The most widely used approach for the shortest distance estimation is the Dijkstra's Algorithm \citep{dijkstra1959note}, which is mainly used to calculate the shortest path. The computational time required by Dijkstra's Algorithm is prohibitively large when dealing with a large roadway network. To reduce computational time, several approaches have been proposed to approximate the shortest distance. For example, the landmark approach \citep{goldberg2005computing} estimates the shortest distance between two locations by approximating it as the minimum of the sum of distances from a source vertex to a set of landmarks and the distances from the landmarks to the destination vertex. It relies on a chosen set of landmarks and their distances to efficiently compute shortest paths.  

Recently, neural networks have been developed to approximate the shortest distance by first embedding node geo-coordinates into an embedding space and then feeding the feature embedding into a feed-forward neural network \citep{qi2020learning,yan2023policy}. However, the generalization capability of these neural network models is not fully investigated. For instance, natural disasters such as earthquakes or hurricanes often result in detours between two locations, leading to a non-negligible increase in the shortest distance. Therefore, a surrogate model relying solely on geo-coordinates can underestimate the actual shortest distance. To address these challenges, we develop a model based on graph neural network (GNNs). GNNs have been successfully used to handle graph data by effectively learning both node and edge representations by extracting neighbor features. The key component of GNNs is to exchange information between nodes and neighbors through message-passing. The generated embedding could be utilized for downstream tasks, including node/edge regression and classification. GNNs have recently been applied to GNNs for regional risk assessment and decision-making \citep{yan2022reinforcement,liu2023optimizing}. In this paper, we propose a novel framework based on GNNs that estimates the shortest distance between two locations in roadway networks and can handle changes in network topologies. We demonstrate the generalization capability of the proposed model in multiple roadway networks. Via a case study, we show how the proposed model can evaluate the flood impact on travel distances on several coastal urban areas.

The outline of the remaining article is as follows. Section \ref{sec:background} provides backgrounds on shortest distance calculation and graph neural network; Section \ref{sec:overview}  presents the framework of GNN-based shortest distance approximation and route recommendation; and Section \ref{sec:experiment} includes the numerical experiments on synthetic and actual transportation networks. 

\section{Technical background}
\label{sec:background}
\subsection{Shortest distance problem}
For the sake of simplicity, we start with a directed, weighted graph, with the set of vertices (or nodes) denoted by  $\bm{V}$ and the set of edges (or links) denoted by $\bm{E}$. The objective in shortest distance problems is to find the shortest travel distance between a given pair of vertices $(s, t)$, with $s$ and $t$ being the origin and destination nodes, respectively. It can be formulated as a linear programming problem:

\begin{equation}
\begin{aligned}
    \min \quad & \sum_{(u,v) \in E} w(u, v) \cdot x_{u,v} \\
    \mathrm{s.t.} \quad & \sum_{(u, v) \in E} x_{u,v} - \sum_{(v, w) \in E} x_{v,w} =
    \begin{cases} 
    1 & \text{if } v = s, \\
    -1 & \text{if } v = t, \\
    0 & \text{otherwise,}
    \end{cases} \\
    & x_{u,v} \in \{0, 1\}, \quad \forall (u, v) \in E.
\end{aligned}
\end{equation}
where $w(u, v)$ is the distance or weight between vertices $u$ and $v$. The binary decision variable $x_{u,v}$ defines whether each edge is selected or not. The flow conservation constraint ensures that the selected edges form a valid path from $s$ and $t$. By solving this linear programming problem, 

\subsection{Graph neural network} 
\label{sec:gnn}
Artificial neural networks (ANNs), as a powerful tool in machine learning, has been applied to a wide range of complex applications such as image segmentation, speech recognition, and predictive modeling. ANNs have the ability to accurately uncover the complex relationships between outputs and inputs by learning the patterns from experimental and simulated data. Without loss of generality, in a feed forward neural networks with multiple layers, at layer $l$, with a $p$-dimensional input vector $\bm{v}^l \in \mathbb{R}^p$, the  $q$-dimensional output $\bm{v}^{l+1} \in \mathbb{R}^q$ is expressed as
\begin{equation}
\label{eq:fnn}
\bm{v}^{l+1} = f(\bm{v}^l; \bm{W}^l, \bm{b}^l) = \sigma(\bm{v}^l \bm{W}^l + \bm{b}^l),
\end{equation}
where $\bm{W}^l \in \mathbb{R}^{p\times q}$ and $\bm{b}^l \in \mathbb{R}^{1\times q}$ denote the learnable weight and bias term, respectively. The function $\sigma(\cdot)$ is a nonlinear activation function. Due to the width limitation of neural network layer and the difficulty of parameter tuning, the capacity and performance of neural networks are  increased by stacking multiple layers.

However, the aforementioned setup of neural networks cannot handle non-Euclidean data, such as graph data, and can only handle input with a fixed size. In order to broaden the application of neural networks to graph data, and to variable-size input belonging to different graph topologies, the graph neural network was proposed. Let us represent a graph  by a four-tuple $\bm{G} =(\bm{V}, \bm{E}, \bm{X}_v, \bm{X}_e)$, where $\bm{X}_v \subset \mathbb{R}^{|\bm{V}|\times F_v}$ and $\bm{X}_e \subset \mathbb{R}^{|\bm{E}|\times F_e}$ represent features of a node $v \in \bm{V}$ and an edge $e \in \bm{E}$, respectively. Also, let $|\bm{V}|$ and $|\bm{E}|$ denote the number of nodes and edges in the graph; and  $F_v$ and $F_e$ denote the number of features for each node and edge, respectively. A key component of a GNN is the ability to model the interdependence between the nodes. This is done by ``message passing", which is the process of exchanging feature attributes (or node embeddings) between nodes along the edge. Multiple message passing approaches have been proposed so for. For instance, \citep{grover2016node2vec} utilized a random walk to find low-dimensional representations for nodes. In another work, the graph convolutional network (GCN) was proposed  to incorporate the adjacency matrix in the finding the embedding of node features \citep{kipf2016semi}. In this method, the forward propagation in $i^{\mathrm{th}}$ layer  is expressed as:

\begin{equation}
\label{eq:gcn}
\bm{h}^{i+1} = \sigma(\widehat{\bm{A}} \bm{h}^i \bm{W}^i),
\end{equation}
where $\sigma$ is a nonlinear activation function, $\bm{W}^i$ and $\bm{h}^i$ denote weight matrix and node embedding at the $i^{\mathrm{th}}$ layer, accordingly. 
$\widehat{\bm{A}}=\tilde{\bm{D}}^{-\frac{1}{2}}\tilde{\bm{A}}\tilde{\bm{D}}^{-\frac{1}{2}}$ is the normalized adjacency matrix and $\tilde{\bm{A}}$ and $\tilde{\bm{D}}$ are the adjacency matrix with a self-loop and degree matrix, respectively.
The adjacency matrix contains the graph connectivity and topology information, allowing the GCN to aggregate node information along the edges. 

\begin{figure}[htbp]
\centering
\includegraphics[width=0.65\textwidth]{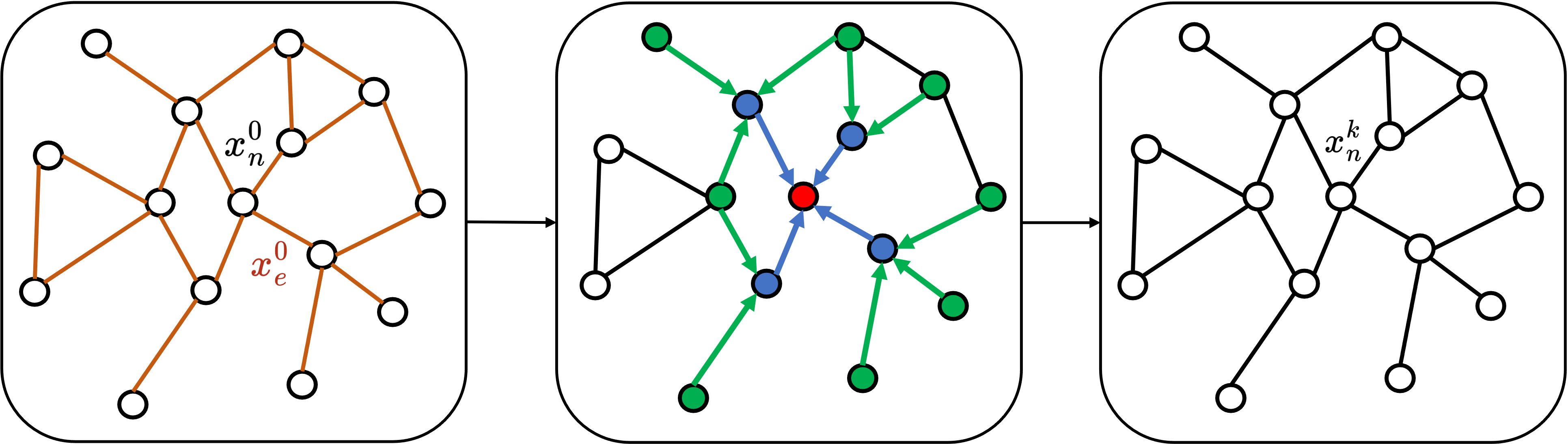}
\caption{Message passing with both node  and edge features}
\label{fig:gnn}
\end{figure}

Another approach for  message passing is the GraphSAGE which  uses generalized neural aggregation functions \citep{hamilton2017inductive}. GraphSAGE learns node embedding by aggregating local neighbor features into  a low-dimensional representation. Figure~\ref{fig:gnn} illustrates the message-passing process with both the node and edge features. The left figure represents the original graph at step 0, where the node features $x^0_n$ and edge feature $x^0_e$ are initialized. Then in the aggregation step, which is shown in the middle figure, the features are passed along the arrow directions. As an example, for the red node, the features from blue nodes and edges are passed into the red node and then updated, which represent 1-hop neighbors. Then in step 2, the node features and edge features with green color are passed into the red node. After a $k$-step update, the node features $x^k_n$ contains information from all $k$-hop neighbors, as exhibited in the right figure. Then the generated node embedding can be further employed for node regression tasks, like single-source shortest distance estimation.

\section{Methodology}
\label{sec:overview}

\begin{figure}[htbp]
\centering
\includegraphics[width=0.8\textwidth]{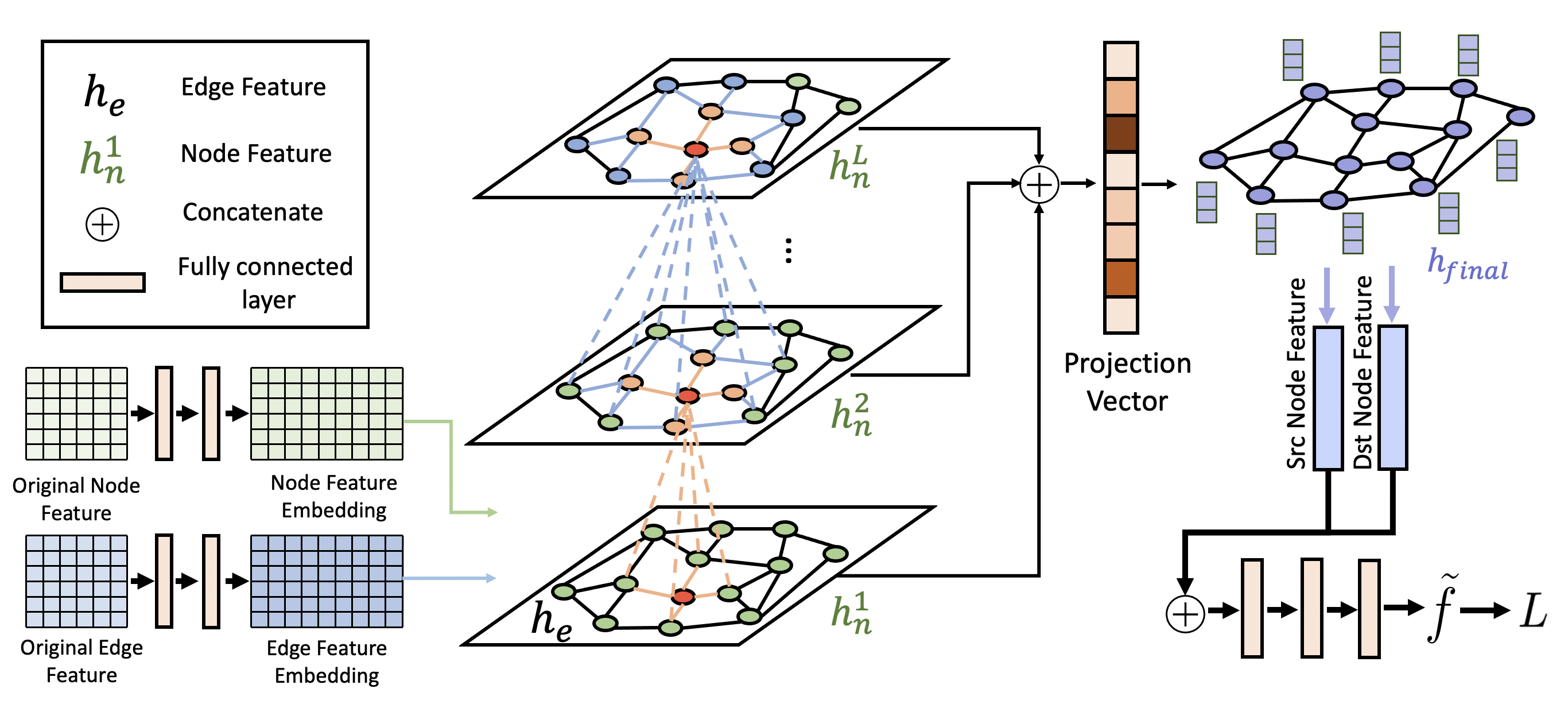}
\caption{The original node and edge features are initially transformed into feature embeddings. These embeddings are then propagated through the GNN layers, where they are concatenated using a projection vector. Subsequently, the features of the source node and destination node are concatenated and passed through a MLP to make the final prediction.}
\label{fig:model}
\end{figure}

In this section, we will introduce the proposed graph neural network framework for shortest distance estimation and shortest path finding. We are motivated by \cite{liu2020towards},  which showed that once the representation transformation is disentangled from the propagation part the performance of GNNs can be improved. Our proposed framework (denoted as GNN-SDE) for the shortest distance prediction is shown in figure \ref{fig:model}. The framework consists of three major stages, including feature pre-processing, feature propagation, and feature prediction. We will elaborate on each component in the following sections.

\subsection{Graph construction and feature selection}

The first step of the pipeline is to build the graph $\bm{G}(\bm{V}, \bm{E})$, where the edges are the road segments in the road network, and nodes are the intersections of the road segments. Furthermore, to achieve better performance for shortest distance estimation, it is critical to use the appropriate set of node features $\bm{X}_v$, and edge features $\bm{X}_e$. The node features $\bm{X}_v \subset \mathbb{R}^{|\bm{V}|\times 3}$ include the number of hop $\mathrm{h}(s, v)$ from source node $s$ to target node $t$ and the coordinates of node $s$ and $t$.  The number of hops indicates the spatial relationship between the source and target nodes. Additionally, in order to create spatially invariant features, the normalized coordinates are adopted in the node feature.   We chose the edge features $\bm{X}_e \subset \mathbb{R}^{|\bm{E}|\times 1}$ to be the weight (travel time or link length) corresponding to each edge. These chosen node and edge features offer a representation of the graph that has information both at local and global levels. 

\subsection{Graph neural network for shortest distance estimation}
\label{sec:distance_estimation}

The shortest distance estimation problem is solved using a node-level regression problem. Before the original features are passed into the GNN propagation, the original node and edge features are transformed into an embedding space using a multi-layer perception (MLP): 
\begin{equation}
\label{eq:embedding}
\begin{aligned}
h^{0}_v &= f(x_v; \bm{W}_v, \bm{b}_v)\\
h_e &= f(x_e; \bm{W}_e, \bm{b}_e), 
\end{aligned}
\end{equation}
where $x_v \in \bm{X}_v$ and $x_e \in \bm{X}_e$ represent original node and edge features and $\bm{W}_v, \bm{b}_v$ and $\bm{W}_e, \bm{b}_e$ denote the learnable parameters for node and edge embedding, respectively. Then, these transformed features, not the original features, are propagated in the message passing. 
The embedding generation of each node are done into two stages: aggregation and update. In the aggregation stage, features from node $\mathcal{N}(v)$ and edges $\mathcal{\bm{E}}(v)$ that are connected to node $v \in \bm{V}$ are aggregated at steps $k = 0, \dots, K-1$, where $K$ is the maximum aggregation step:


\begin{equation}
\label{eq:node_aggr}
\begin{aligned}
h^{k+1}_{\mathcal{N}(v)} &= \mathrm{AGGREGATE}_k(\{h_u^k, \forall u \in \mathcal{N}(v)\})\\
h^{k+1}_{\mathcal{\bm{E}}(v)} &= \mathrm{AGGREGATE}_k(\{h_e^k, \forall e \in \mathcal{\bm{E}}(v)\}),
\end{aligned}
\end{equation}
where $\mathrm{AGGREGATE}_k$ is a function aggregating the embeddings of a node's local neighborhood at layer $k$. 
Then in the update stage, the updated node feature $h_v^{k+1}$ in step $k+1$ will be calculated using the following feed-forward network:

\begin{equation}
\label{eq:node_update}
\begin{aligned}
h^{k+1}_{v} &= f\left( \mathrm{CONCAT} \left(\{h^k_v, h^{k+1}_{\mathcal{N}(v)}, h^{k+1}_{\mathcal{\bm{E}}(v)}\}\right ); \bm{W}^{k}, \bm{b}^{k} \right),
\end{aligned}
\end{equation}
where $\mathrm{CONCAT}$ is the concatenation function; $\bm{W}^{k}$ and $\bm{b}^{k}$ are the learnable parameters in the feed-forward network. After the $K$-step propagation, at the central node, the new node embedding is generated using the convolution of the $K$-hop node and edge features. Finally, the node embeddings from all layers are concatenated together and multiplied by a trainable projection vector to calculate the final node embedding:
\begin{equation}
\label{eq:final_update}
\begin{aligned}
h_{final} &=  f \left(\mathrm{CONCAT} \left(\{h^0_v, h^1_v, \dots, h^K_v\}\right ) \bm{s}\right) \\
y_{final} &=  \mathrm{MLP} \left( h_{final}; \bm{W}_d, \bm{W}_{d} \right ), 
\end{aligned}
\end{equation}
where $\bm{s} \in \mathbb{R}^{(K+1) \times 1}$ is a trainable projection vector. The projection vector has similar functionality as residual connections to facilitate better information flow throughout the network. Furthermore, it allows for a more flexible representation of each node and preserves the information from more layers. The final embedding $h_{final}$ is passed into the prediction blocks with a multi-layer perceptron to make predictions, which predicts the shortest distance $y_{final} \subset \mathbb{R}^{|\bm{V}|\times 1}$ from a single source to all nodes. 

The weighted mean absolute error is chosen as the loss function and minimized in the backward propagation. The weight $w \in \mathbb{R}^{|\bm{V}|\times 1}$ for each node $v \in \bm{V}$ is chosen as the reciprocal of the shortest distance for node $v$ from the source node. The weights are further heuristically clamped into the range from 0.1 to 1.0. Furthermore, It is computationally expensive for the model to involve all nodes in the loss function and backpropagation. In order to shorten the training process, a binary mask $m \in \mathbb{R}^{|\bm{V}|\times 1}$ is applied to the graph, which specifically targets a subset of nodes for the loss function instead of considering all the nodes. The loss function $L$ can be expressed as follows:

\begin{equation}
L = \frac{1}{B}\sum_{i=1}^{B} {\left \| m_i \otimes w_i \otimes (y_{s, i} - \tilde{y}_{s, i}) \right \|}_1, 
\end{equation}
where $B$ is the batch size and $\otimes$ is the element-wise multiplication. The ground truth data and GNN prediction in the dataset are denoted as the shortest distance $y_s \in \mathbb{R}^{|\bm{V}|\times 1}$ and $\tilde{y}_s \in \mathbb{R}^{|\bm{V}|\times 1}$ from source node $s$ of $i^{\mathrm{th}}$ sample from the batch, which is calculated using the Dijkstra's algorithm and GNN, respectively.

\subsection{Shortest Path Finding and Route Recommendation}
\label{sec:path_finding}

As discussed in Section \ref{sec:gnn}, while GNN models are well-suited for node-level or edge-level predictions, they inherently face challenges in generating route recommendations directly. However, inspired by the Dijkstra algorithm, we can transform the shortest path-finding problem into a problem of identifying the predecessor of each node, thereby constructing paths from a single source to all other nodes. This pathfinding process serves as a subsequent step following the shortest distance estimation in Section \ref{sec:distance_estimation}. The node-level shortest distance estimation discussed in Section \ref{sec:distance_estimation} provides the foundation for this process. For each node in the graph, the path to the source node can be recursively reconstructed by tracing back through its predecessors. The detailed implementation of the pathfinding algorithm is presented in Algorithm \ref{alg:path_finding}.

\begin{algorithm}[H]
\label{alg:path_finding}
\SetAlgoLined
\KwIn{Road network $G = (V, E)$, source node $s$, target node $t$, distance estimation from source node $\mathcal{D}$}
\KwOut{Shortest path from $s$ to $t$}

\textbf{Step 1: Compute Predecessors}\;
Initialize an empty dictionary $predecessors\_dict$\;
\ForEach{$v \in V$}{
    \If{$v == s$}{
        \textbf{continue}\;
    }
    Initialize $res \leftarrow \infty$, $predecessor \leftarrow \text{None}$\;
    \ForEach{neighbor $u$ of $v$}{
        \If{$dist\_dict[u] < res$}{
            $predecessor \leftarrow u$\;
            $res \leftarrow dist\_dict[u]$\;
        }
    }
    $predecessors\_dict[v] \leftarrow predecessor$\;
}

\textbf{Step 2: Construct the Path}\;
Initialize an empty list $path$\;
Set $current \leftarrow t$\;

\While{$current \neq s$}{
    Append $current$ to $path$\;
    $current \leftarrow predecessors\_dict[current]$\;
}
Append $s$ to $path$\;
\Return $path$\;
\caption{Shortest Path Finding and Route Recommendation}
\end{algorithm}

\section{Numerical experiment}
\label{sec:experiment}

Two sets of numerical experiments are conducted to evaluate the feasibility and efficiency of the proposed GNN-based shortest distance algorithm. The first experiment uses synthetic graphs with randomized edge length to demonstrate the feasibility and generalization capability of the proposed model. The second experiment uses real coastal urban networks to showcase the capabilities of the GNN model for use in urban planning, specifically in the planning of efficient shelter system and evacuation, to mitigate flood impacts. 

\subsection{Synthetic networks with random parameters}
The synthetic roadway networks are generated based on  grid graphs with random node locations and hypotenuse links between nodes. Starting from an initial grid,  some of the nodes and edges are randomly removed to emulate  real world scenarios such as road closures. The edge weights represent the distance between nodes. Different graph sizes studied in this work include graphs with 1k, 2k, 3k, 4k, 10k, 20k, 50k, and 100k nodes. 
In order to calculate the shortest distance from a source node, each time the source node changes, a new graph should be formed, since some of the node features depend on the target node. The synthetic roadmap dataset is then separated into the training set $\mathcal{D}_{train}$ and testing set $\mathcal{D}_{test}$. The graph topologies in $\mathcal{D}_{train}$ and $\mathcal{D}_{test}$ are different to demonstrate the generalization capability of the proposed model. Separate models were trained and tested for each graph size.

\begin{table}[!htb]
\centering
\caption{Training Setting with synthetic dataset}
\label{tab:training_setting}
\resizebox{0.6\textwidth}{!}{%
\begin{tabular}{ccccccccc}
\hline
Graph Size    & 1k   & 2k   & 3k   & 4k   & 10k  & 20k  & 50k  & 100k \\ \hline
\# topology   & 10   & 10   & 10   & 10   & 5    & 5    & 5    & 5    \\ \hline
$|\mathcal{D}_{train}|$ & 6400 & 6400 & 3200 & 3200 & 2500 & 2500 & 2500 & 2500 \\ \hline
$|\mathcal{D}_{test}|$  & 1600 & 1600 & 800  & 800  & 500  & 500  & 500  & 500  \\ \hline
Batch Size    & 64   & 64   & 32   & 32   & 16   & 16    & 4    & 4    \\ \hline
\end{tabular}%
}
\end{table}

\begin{figure}[!hbt]
\centering
\begin{subfigure}[]{0.25\textwidth}
 \centering
 \includegraphics[width=\textwidth]{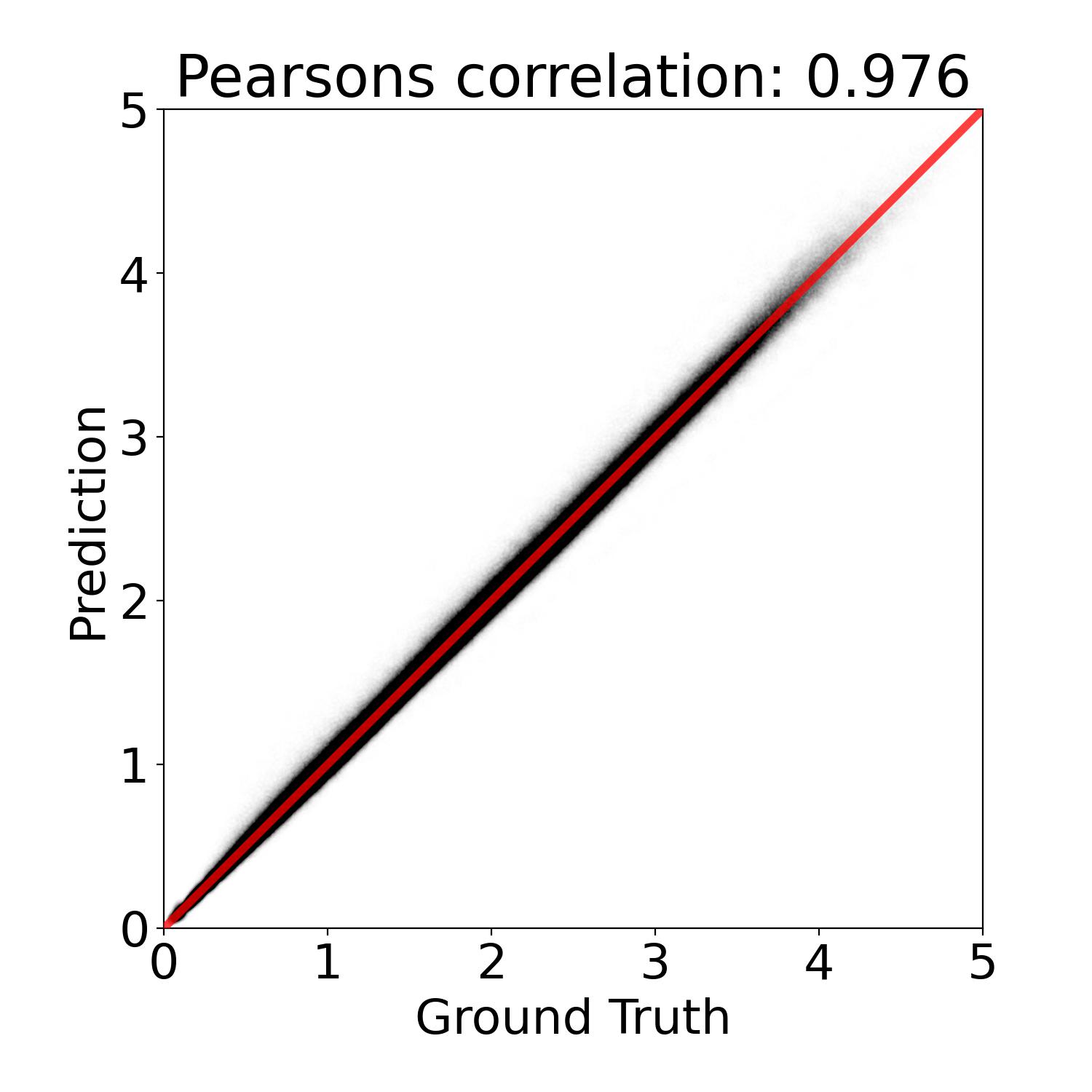}
 \caption{graph size 1k}
 \label{fig:corr_300}
\end{subfigure}
\quad
\begin{subfigure}[]{0.25\textwidth}
 \centering
 \includegraphics[width=\textwidth]{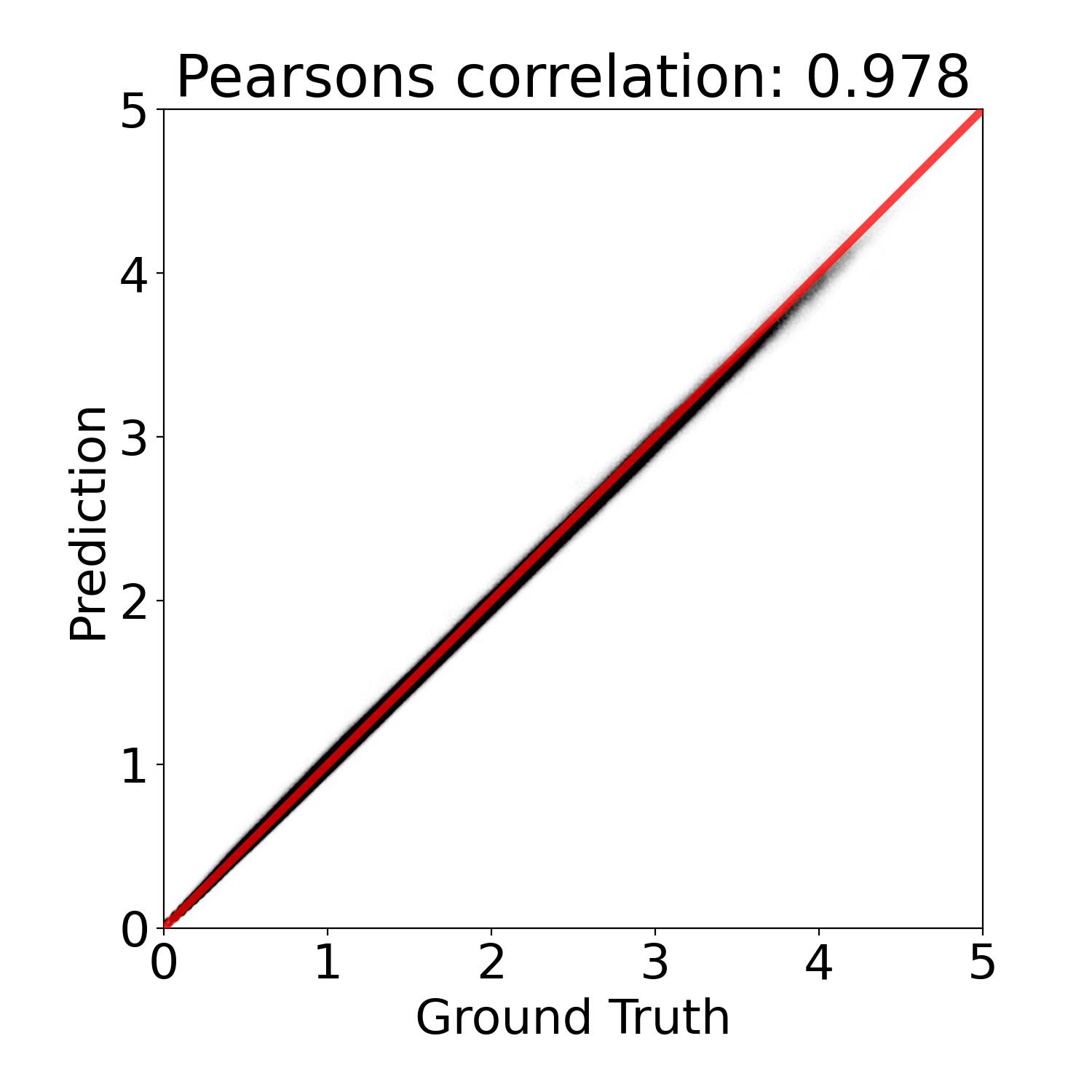}
 \caption{graph size 4k}
 \label{fig:corr_4k}
\end{subfigure}
\quad
\begin{subfigure}[]{0.25\textwidth}
 \centering
 \includegraphics[width=\textwidth]{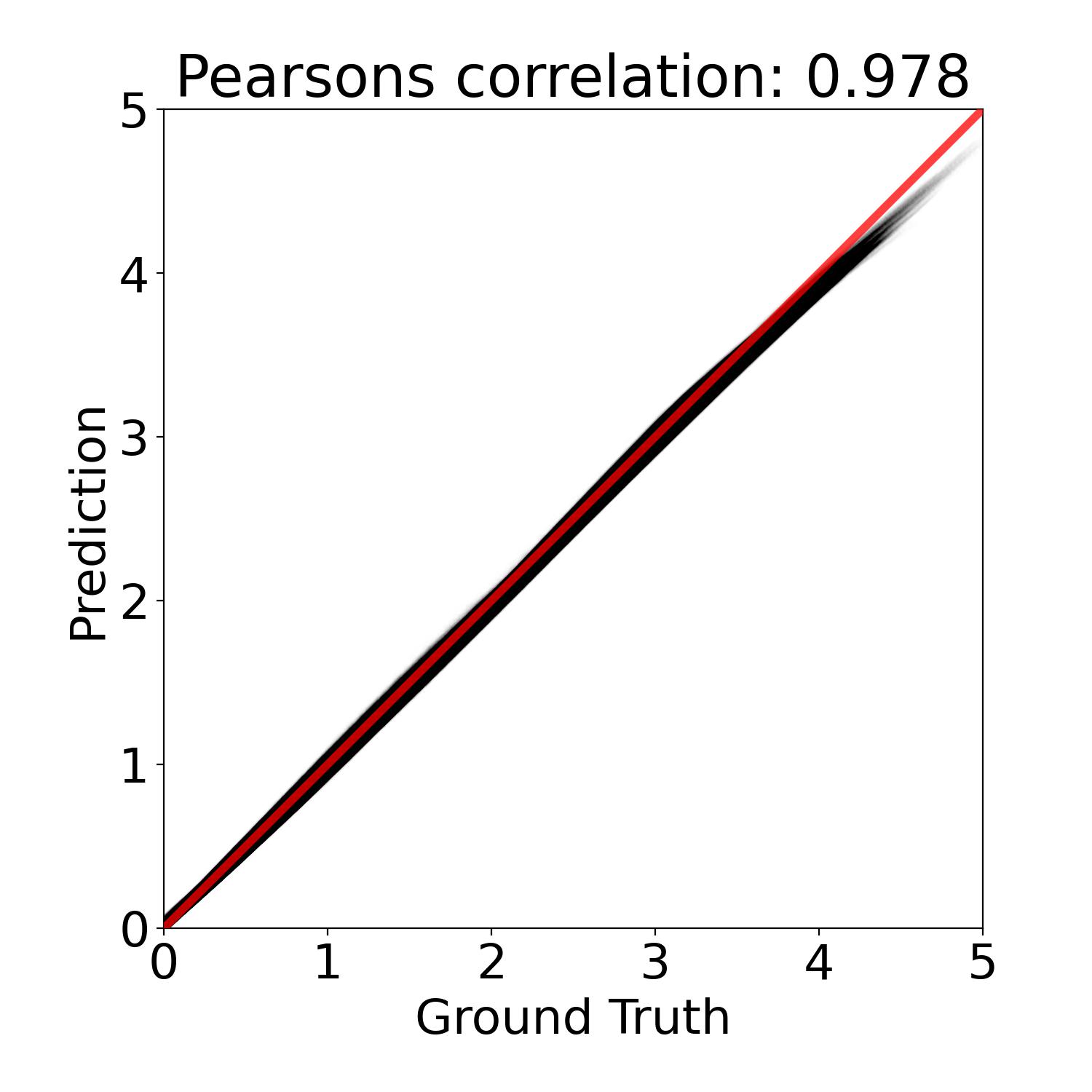}
 \caption{graph size 50k}
 \label{fig:corr_50k}
\end{subfigure}
\caption{Shortest distance prediction under different graph sizes. The prediction from GNN-SDE is compared with Dijkstra's Algorithm under three different graph size, including 1k, 4k, and 50k.}
\label{fig:corr}
\end{figure}

\begin{figure}[htb]
     \centering
     \begin{subfigure}[htb!]{0.4\textwidth}
         \centering
         \includegraphics[width=\textwidth]{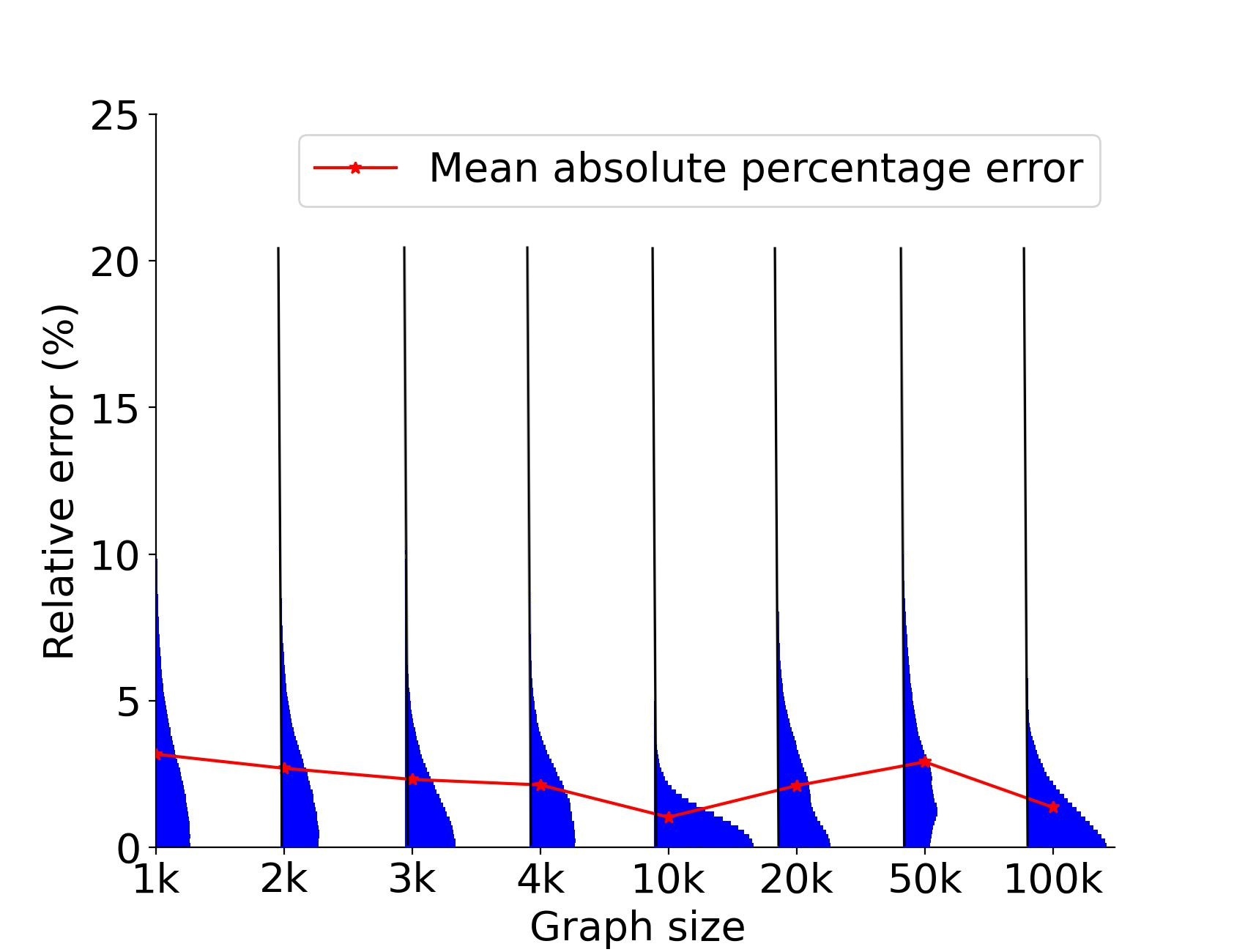}
         \caption{Absolute percentage error distribution}
         \label{fig:Rerr_vert}
     \end{subfigure}
     \begin{subfigure}[htb!]{0.4\textwidth}
         \centering
         \includegraphics[width=\textwidth]{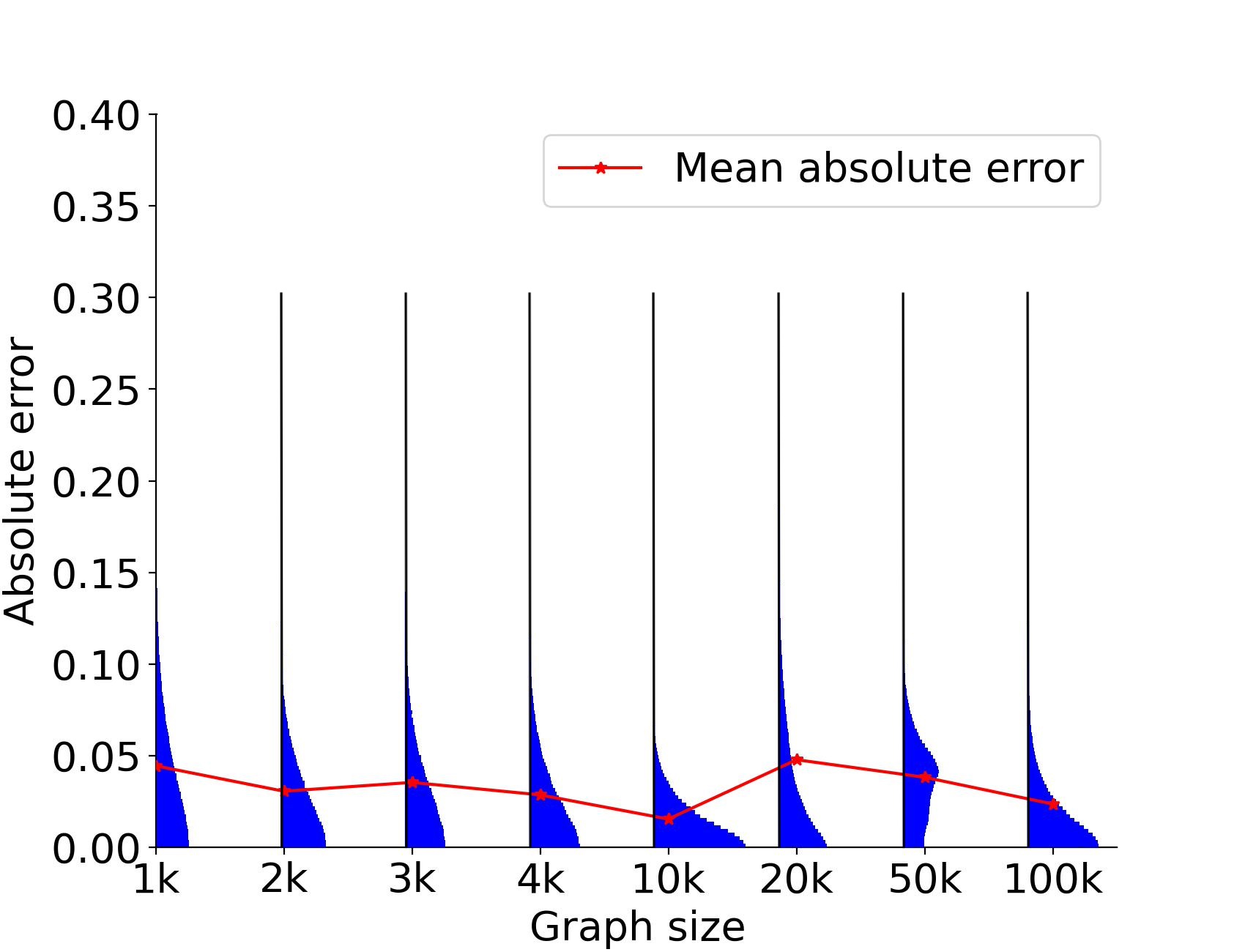}
         \caption{Absolute error distribution}
         \label{fig:Aerr_vert}
     \end{subfigure}
     \hfill
        \caption{Relationship between error distribution versus different graph sizes. The blue distributions are error histograms tilted by 90 degrees. The red line connects the mean absolute percentage errors (MAPE) and mean absolute errors (MAE), respectively, between graph sizes. The taller blue distribution refer to wider range of errors. The peak of error histograms are mostly at the lowest-error bin.}
        \label{fig:err_vert}
\end{figure}

The detailed training setting of each dataset including batch size, training/testing size, and number of topologies is shown in Table \ref{tab:training_setting}. The GNN training is implemented using PyTorch \citep{paszke2019pytorch}. The transformation block and prediction block are chosen as a three-layer fully connected network. And three GNN layers are used between the transformation block and prediction block, as described in Section \ref{sec:overview}. For hyper-parameter selection, the hidden layer size is chosen as 64, which is common in neural network implementation. Mini-batch stochastic gradient descent was implemented in the training process using adaptive moment estimation optimizer (Adam) \citep{kingma2014adam} with a learning rate of 0.001 to find the optimal parameters. The training epoch is set as 100. To evaluate the model performance, mean absolute error (MAE) and mean absolute percentage error (MAPE), given by the following expressions, are compared.

\begin{equation}
\begin{aligned}
    \label{eq:metric}
    \mathrm{MAE} &= \frac{1}{|\mathcal{D}_{test}|} \sum_{(s,t) \in \mathcal{D}_{test}} |y_{s,t} - \tilde{y}_{s,t}|, \\
    \mathrm{MAPE} &= \frac{1}{|\mathcal{D}_{test}|}\sum_{(s,t) \in \mathcal{D}_{test}} \frac{|y_{s,t} - \tilde{y}_{s,t}|}{y_{s,t}},
\end{aligned}
\end{equation}
where the $y_{s,t}$ and $\tilde{y}_{s,t}$ represents the ground truth and the GNN-based prediction of shortest distances between node pair $(s,t)$ in the testing dataset $\mathcal{D}_{test}$. The error distributions calculated over different graphs sizes are shown in Figures~\ref{fig:Rerr_vert} and \ref{fig:Aerr_vert}.  These figures included tilted histograms of the MAPE and MAE values for each testing dataset.  As can be seen in Figure \ref{fig:err_vert}, most of the cases are concentrated in the lower-error bins, while only a small portion of the cases exhibits large error. The mean relative error of GNN prediction is less than 2\% when the graph size varies from 1k to 100k. Additionally, the MAPE distribution of the larger graph has less variance compared to smaller graphs. To measure the statistical difference between the GNN-based shortest distance predictions and the ground truth, we use the Pearson's correlation coefficient given by:

\begin{equation}
    \gamma = \frac{\sum (x_i-\bar{x})(y_i-\bar{y})}{\sqrt{\sum (x_i-\bar{x})^2\sum (y_i-\bar{y})^2}},
    \label{eq:pearson}
\end{equation}
where ${\displaystyle \left\{(x_{1},y_{1}),\ldots ,(x_{n},y_{n})\right\}}{\displaystyle \left\{(x_{1},y_{1}),\ldots ,(x_{n},y_{n})\right\}}$ is the $n$ data pairs need to measured. Correlation plots between the GNN prediction and ground truth at graph sizes of 1k, 4k, and 50k are shown in Figure \ref{fig:corr}, revealing the Pearson correlation coefficients of 0.979, 0.98, and 0.979, respectively. This indicates that the GNN-SDE model offers  a good approximation  of the exact results obtained  from the Dijkstra's Algorithm.

Furthermore, the proposed GNN model is compared with several baseline methods: landmark method \citep{beinhofer2011near} and node2vec \citep{grover2016node2vec}. Two variants of node2vec are used, namely  node2vec-sub and node2vec-cat. The main difference between these two variants is that node embeddings are utilized differently by subtracting and concatenating the source and destination node embeddings, respectively. The number of landmarks is chosen as $2\%|\bm{V}|$ for graph size less than 10k, and $0.5\%|\bm{V}|$ for graph size larger than 10k. The training settings for node2vec are adopted from \citep{grover2016node2vec} with the embedding size of 128. We compare the MAPE, training time, and inference time among different approaches. The training time includes the time for preprocessing (landmark, node2vec) and neural network training (node2vec, GNN). The inference time refers to the time for predicting the shortest distance and calculating the shortest path. The performance comparison with graph sizes of 2k, 10k, and 20k is shown in Table \ref{tab:comparison_baseline}. The landmark approach requires minimal time for preprocessing but the inference time is the longest. Node2vec method reduces the inference time but takes longer to generate node embeddings as graph size increases, and performs worse than the landmark approach and GNN model on large graphs. However, the proposed GNN model achieves high accuracy with a relatively small training time by learning the node and edge features and preserving them in the node embeddings.

\begin{table}[!htb]
\renewcommand{\arraystretch}{1.2}
\centering
\caption{Performance Comparison between the proposed GNN model with baseline approaches}
\label{tab:comparison_baseline}
\resizebox{\textwidth}{!}{
\begin{tabular}{c|cccc|cccc|cccc}
\hline
\multirow{2}{*}{Method} &
  \multicolumn{4}{c}{N = 2000} &
  \multicolumn{4}{c}{N = 10000} &
  \multicolumn{4}{c}{N = 20000} \\ \cline{2-13}
 &
  \splitcell{Training \\ time (s)} &
  \splitcell{Inference \\ time (s)} &
  MAE &
  MAPE &
  \splitcell{Training \\ time (s)} &
  \splitcell{Inference \\ time (s)} &
  MAE &
  MAPE &
  \splitcell{Training \\ time (s)} &
  \splitcell{Inference \\ time (s)} &
  MAE &
  MAPE \\ \hline
landmark     & 1.81     & 79.20 & 0.175 & 8.46\%  & 7.46      & 240.00 & 0.175 & 8.46\% & 22.52     & 317.30 & 0.095 & 4.14\%  \\
node2vec-sub & 969.07   & 4.42  & 0.247 & 12.61\% & 1492.87   & 10.38  & 0.199 & 9.61\% & 1881.42   & 19.40   & 0.262 & 12.41\% \\
node2vec-cat & 670.50   & 4.44  & 0.188 & 9.12\%  & 1568.32   & 11.14  & 0.201 & 9.96\% & 1812.99   & 21.52   & 0.194 & 8.59\%  \\
GNN-SDE      & 275.74   & 5.11  & 0.045 & 2.24\%  & 495.05    & 14.69  & 0.030 & 1.45\% & 564.36    & 15.32   & 0.058 & 2.49\% \\ \hline
\end{tabular}
}
\end{table}

\begin{figure}[htb]
    \centering
     \includegraphics[width=0.4\textwidth]{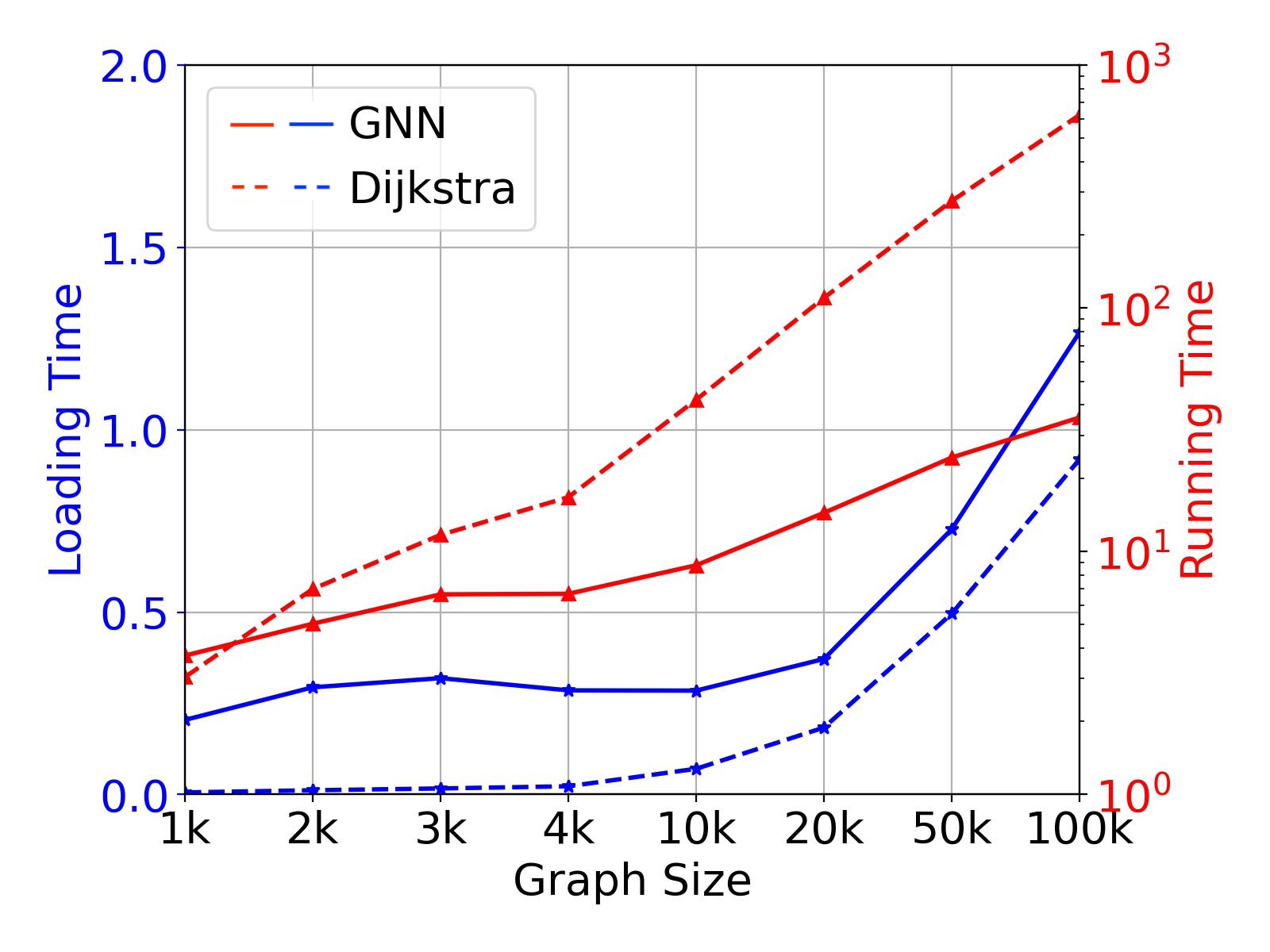}
     \caption{Computational time comparison between graph neural network and Dijkstra}
     \label{fig:time}
\end{figure}

A comparison of computational efficiency between the  GNN model and Dijkstra's algorithm is also conducted. The computational time for both approaches consists of two parts: loading time and running time, which respectively refer to the time to initialize the graph, and the inference time used to calculate the shortest distance and shortest path between node pairs. 1000 evaluations of each graph size were conducted using both methods. The efficiency comparison is shown in Figure \ref{fig:time}. The loading time of GNN is relatively higher than that of the Dijkstra's method, where the time difference is attributed to the node and edge feature generation and computational overhead. When running time is concerned,  the GNN model is more efficient than the Dijkstra's algorithm as the graph size grows, reaching a 10x speed up for the 20k-node graph.

\subsection{Coastal urban network}
\label{sec:casestudy}

Two coastal networks, including Manhattan, New York, and Hillsborough County, Florida, are selected to demonstrate how the proposed model can measure the impact of flood hazards on individual evacuation times at different locations of coastal metropolitan cities, and enhance the disaster planning. The U.S. Street Network dataset \citep{usstreet} is used to construct the roadmap network. The network consists of nodes that correspond to road intersections and edges that represent the roadways, with the length of each edge indicating the distance between the connected nodes.

\begin{figure}[htb!]
\centering
\begin{minipage}{0.49\textwidth}
\begin{subfigure}[htb!]{0.49\linewidth}
 \centering
 \includegraphics[width=\linewidth]{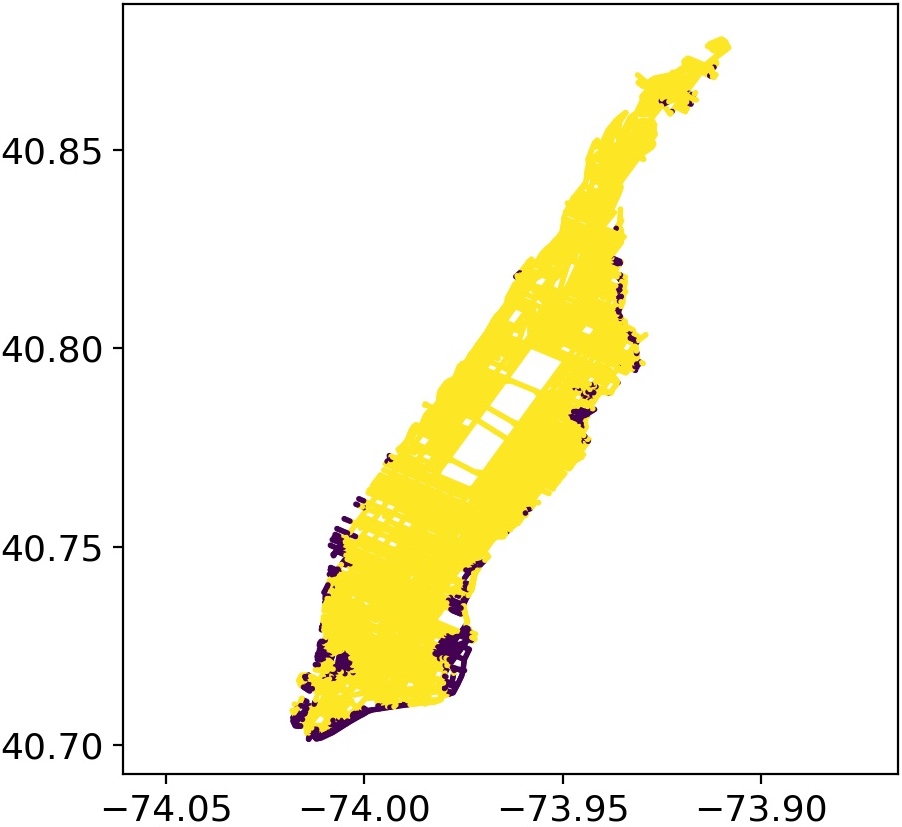}
 \caption{Manhattan}
 \label{fig:manhattan_flood_1}
\end{subfigure}
\begin{subfigure}[htb!]{0.49\linewidth}
 \centering
 \includegraphics[width=\linewidth]{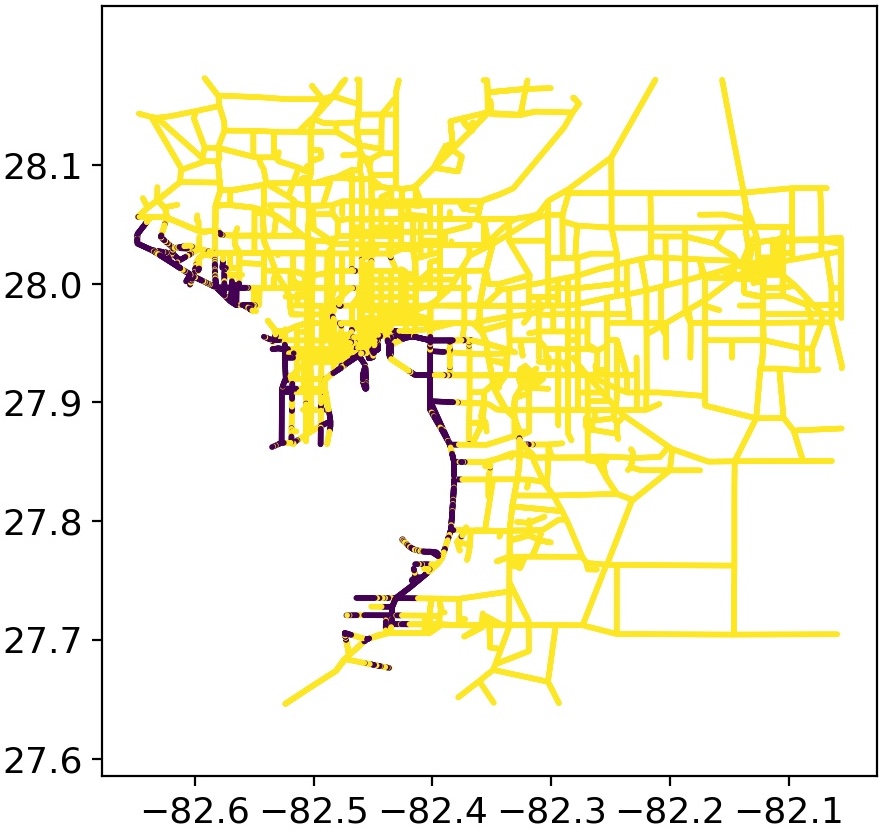}
 \caption{Hillsborough County}
 \label{fig:hillsborough_flood_1}
\end{subfigure}
\caption{Flooded area for each region under Category 1}
\label{fig:flood_1}
\end{minipage}
\begin{minipage}{0.49\textwidth}
\begin{subfigure}[htb!]{0.49\linewidth}
 \centering
 \includegraphics[width=\linewidth]{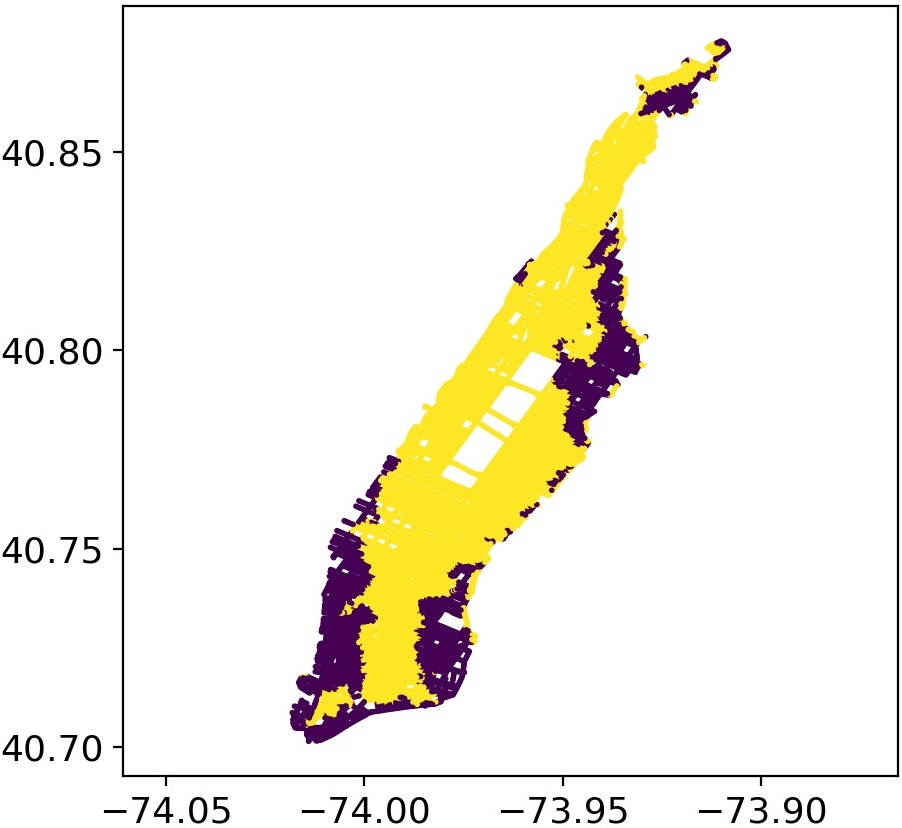}
 \caption{Manhattan}
 \label{fig:manhattan_flood_3}
\end{subfigure}
\begin{subfigure}[htb!]{0.49\linewidth}
 \centering
 \includegraphics[width=\linewidth]{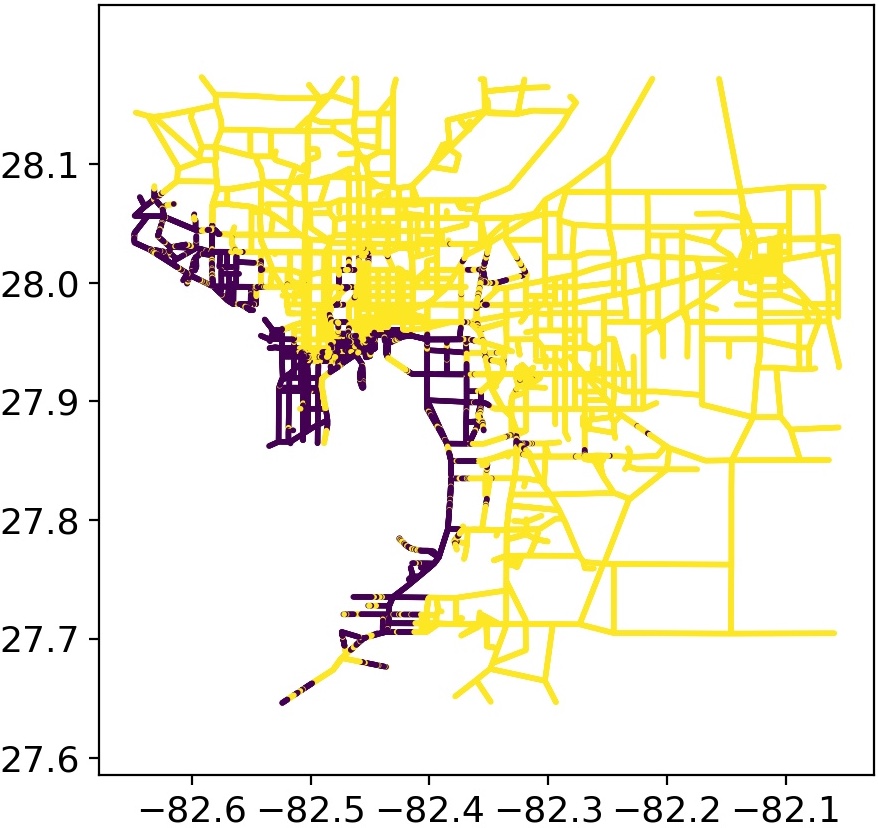}
 \caption{Hillsborough County}
 \label{fig:hillsborough_flood_3}
\end{subfigure}
\caption{Flooded area for each region under Category 3}
\label{fig:flood_3}
\end{minipage}
\end{figure}

During a flood, the ability to quickly identify the most efficient routes for emergency services, evacuation efforts, and resource distribution can significantly reduce response times and save lives. The shortest distance and shortest time estimation help optimize the movement of personnel and resources, ensuring they reach affected areas with minimal delay. Furthermore, it aids in decision-making by providing essential data for route planning, risk assessment, and the prioritization of high-risk zones, ultimately improving the effectiveness and coordination of emergency response efforts. 

The performance of the network is evaluated following a flooding event in the area. To quantify the risk and impact of flood hazards to metropolitan areas, the SLOSH model \citep{jelesnianski1984slosh} is used. Maximum envelopes of water (MEOWs) are generated using 100,000 hypothetical storms with five intensity levels ranging from Category 1 to Category 5. The resulting flood data includes an inundation map of the study area. The region is considered as flooded area when the indentation level is higher than 1ft, which reaches the stability limit for small passenger vehicles \citep{shah2021review}. The flooded impacts at each region under the Category 1 and 3 hazards are shown in Figure \ref{fig:flood_1} and \ref{fig:flood_3}, respectively. It is observed that the impact of Category 3 hurricanes is more widespread than Category 1, resulting in more extensive damage to the infrastructure system.

\begin{figure}[htb!]
\centering
\begin{minipage}[t]{0.7\textwidth}
\begin{subfigure}[htb!]{0.49\linewidth}
 \centering
 \includegraphics[width=\linewidth]{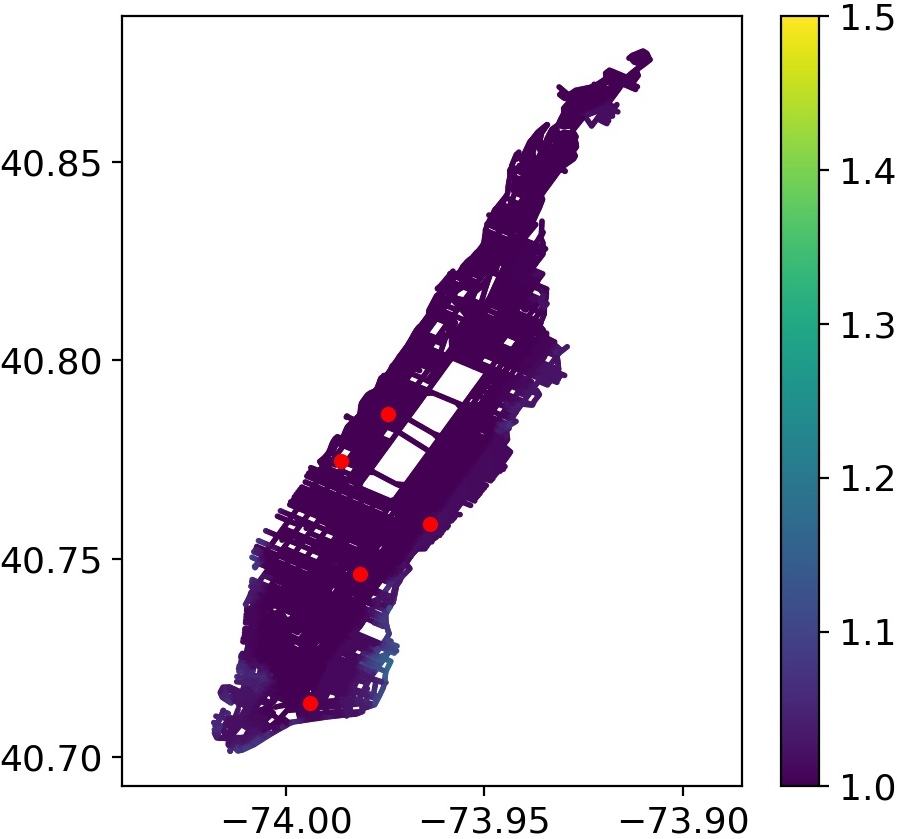}
 \caption{Category 1}
 \label{fig:manhattan_1}
\end{subfigure}
\hfill
\begin{subfigure}[htb!]{0.49\linewidth}
 \centering
 \includegraphics[width=\linewidth]{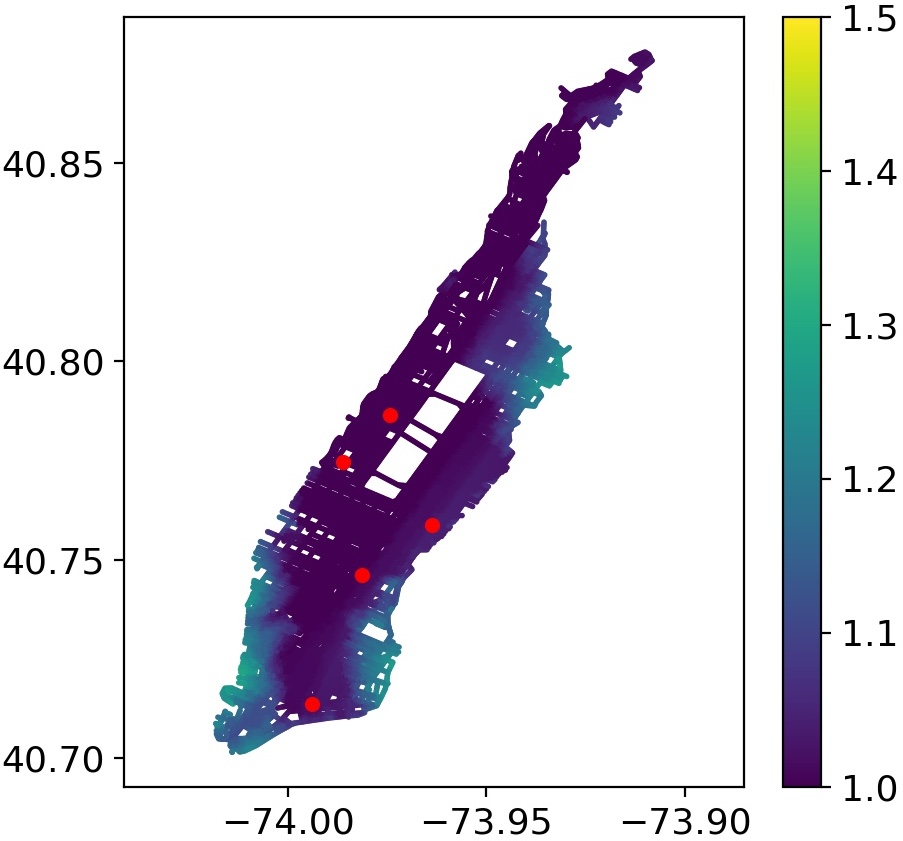}
 \caption{Category 3}
 \label{fig:manhattan_3}
\end{subfigure}
\caption{Delay ratios at Manhattan under different categories. The red dots represent the locations of public evacuation shelters.}
\label{fig:time_1}
\end{minipage}

\begin{minipage}[t]{0.7\textwidth}
\begin{subfigure}[htb!]{0.49\linewidth}
 \centering
 \includegraphics[width=\linewidth]{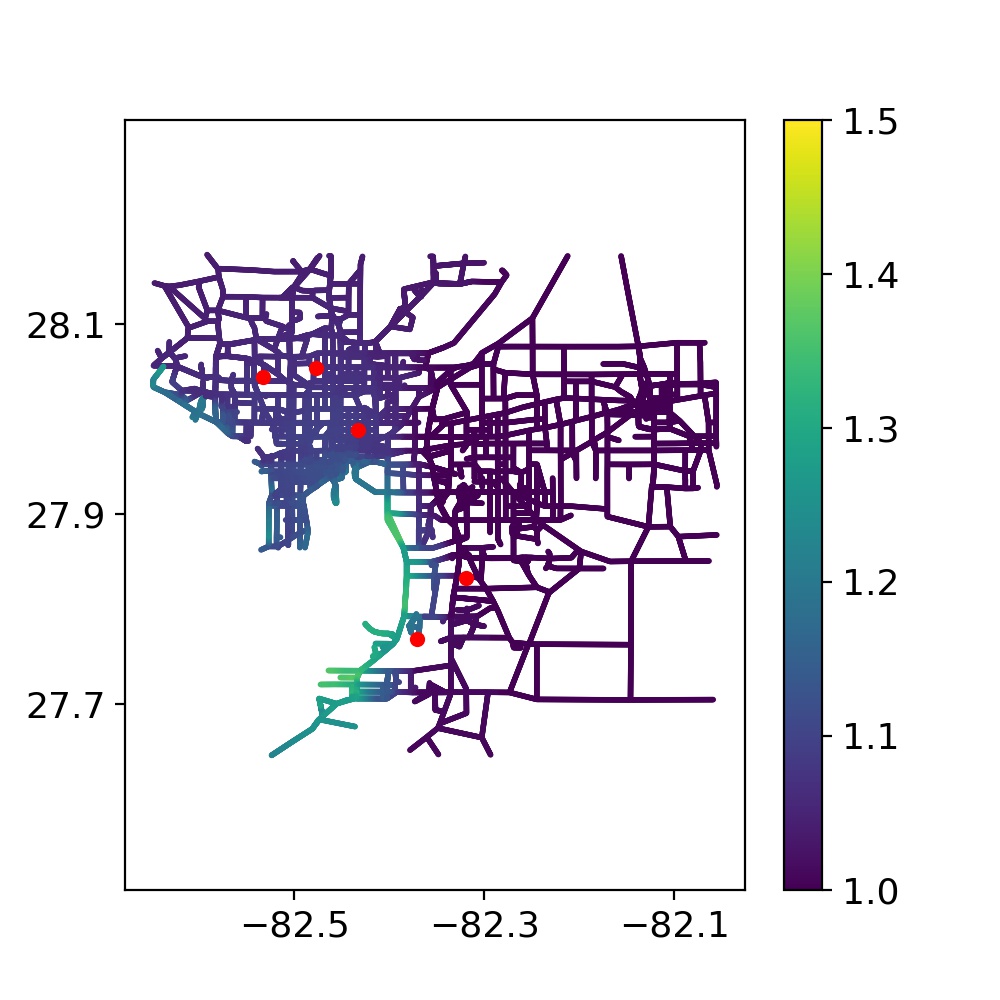}
 \caption{Category 1}
 \label{fig:hillsborough_1}
\end{subfigure}
\hfill
\begin{subfigure}[htb!]{0.49\linewidth}
 \centering
 \includegraphics[width=\linewidth]{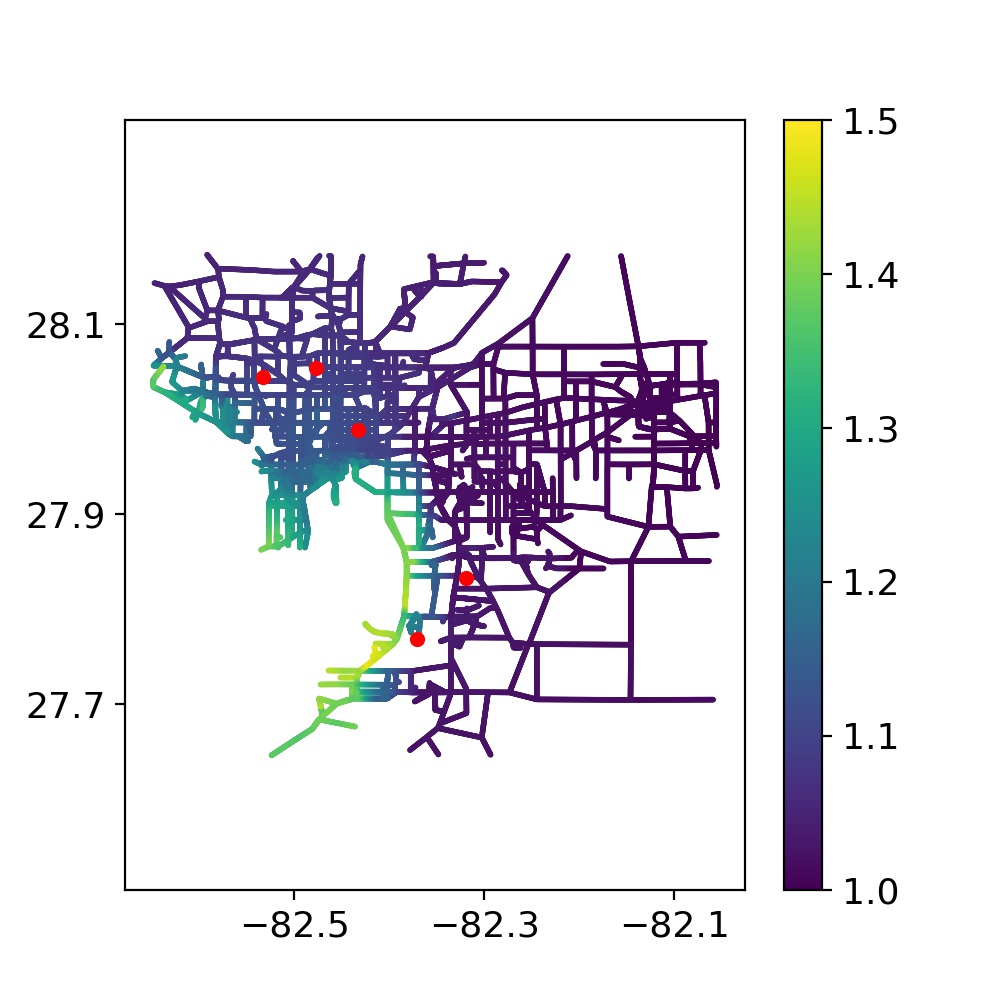}
 \caption{Category 3}
 \label{fig:hillsborough_3}
\end{subfigure}
\caption{Delay ratios at Hillsborough County under different categories. The red dots represent the locations of public evacuation shelters.}
\label{fig:time_3}
\end{minipage}
\end{figure}

We then measure the impact of flood hazards on individual times needed for evacuation to public shelters. To this end, the output of the GNN model is $t_0$ and $t_1$, which are the shortest evacuation times of a given location to the closest public shelter, before and after the flood, respectively. The delay ratio $\delta=t_0/t_1$ is then used as a measure of the flood impact on the evacuation time delay for local residents. Furthermore, the public shelters, gathered from the government websites \citep{manhanttan, hillsborough}, are set to be the source nodes in each investigated network.

\begin{figure}[htb!]
     \centering
     \begin{subfigure}[htb!]{0.4\textwidth}
         \centering
         \includegraphics[width=\textwidth]{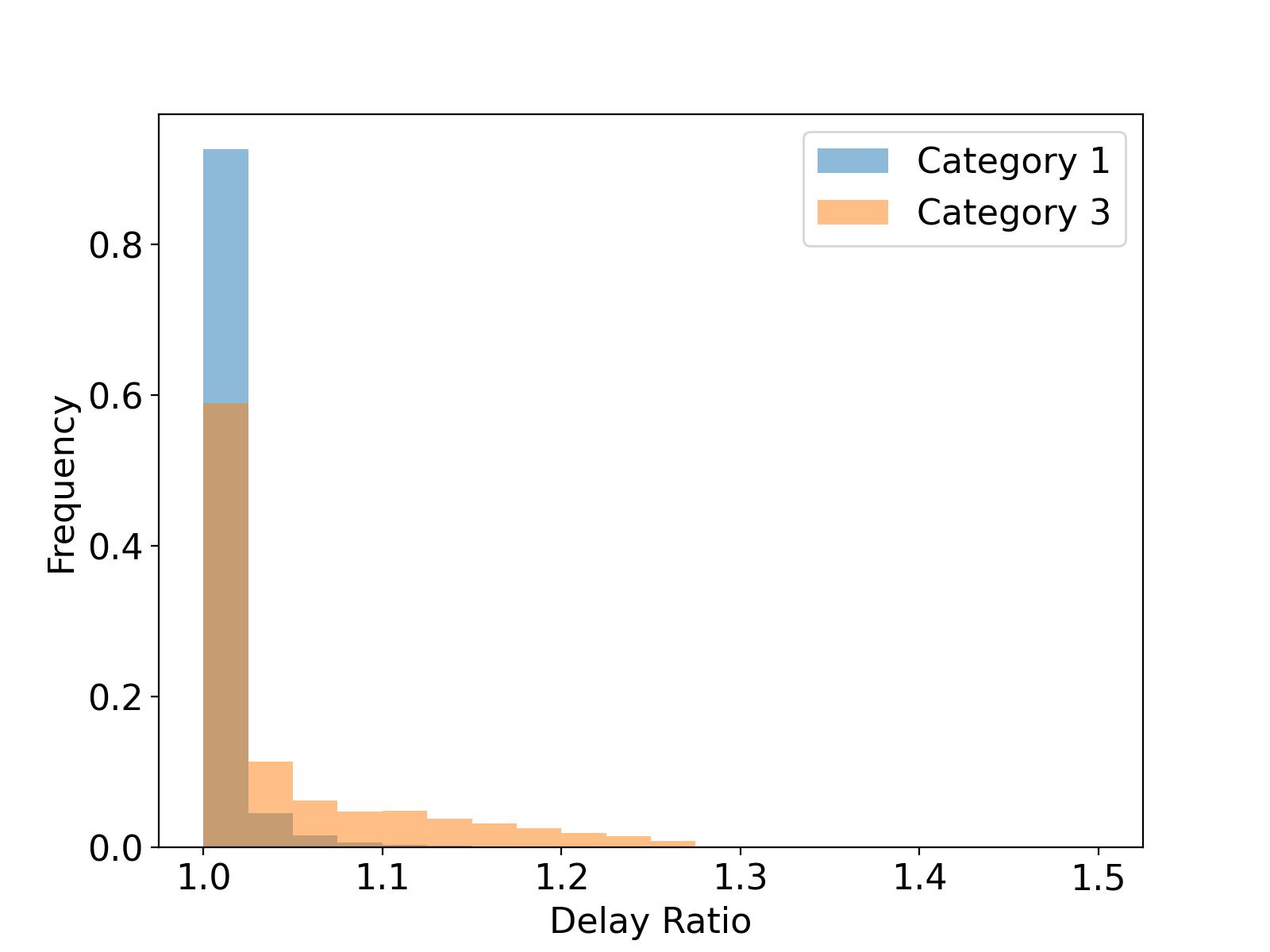}
         \caption{Manhattan}
         \label{fig:hist_manhattan_1_3}
     \end{subfigure}
     \begin{subfigure}[htb!]{0.4\textwidth}
         \centering
         \includegraphics[width=\textwidth]{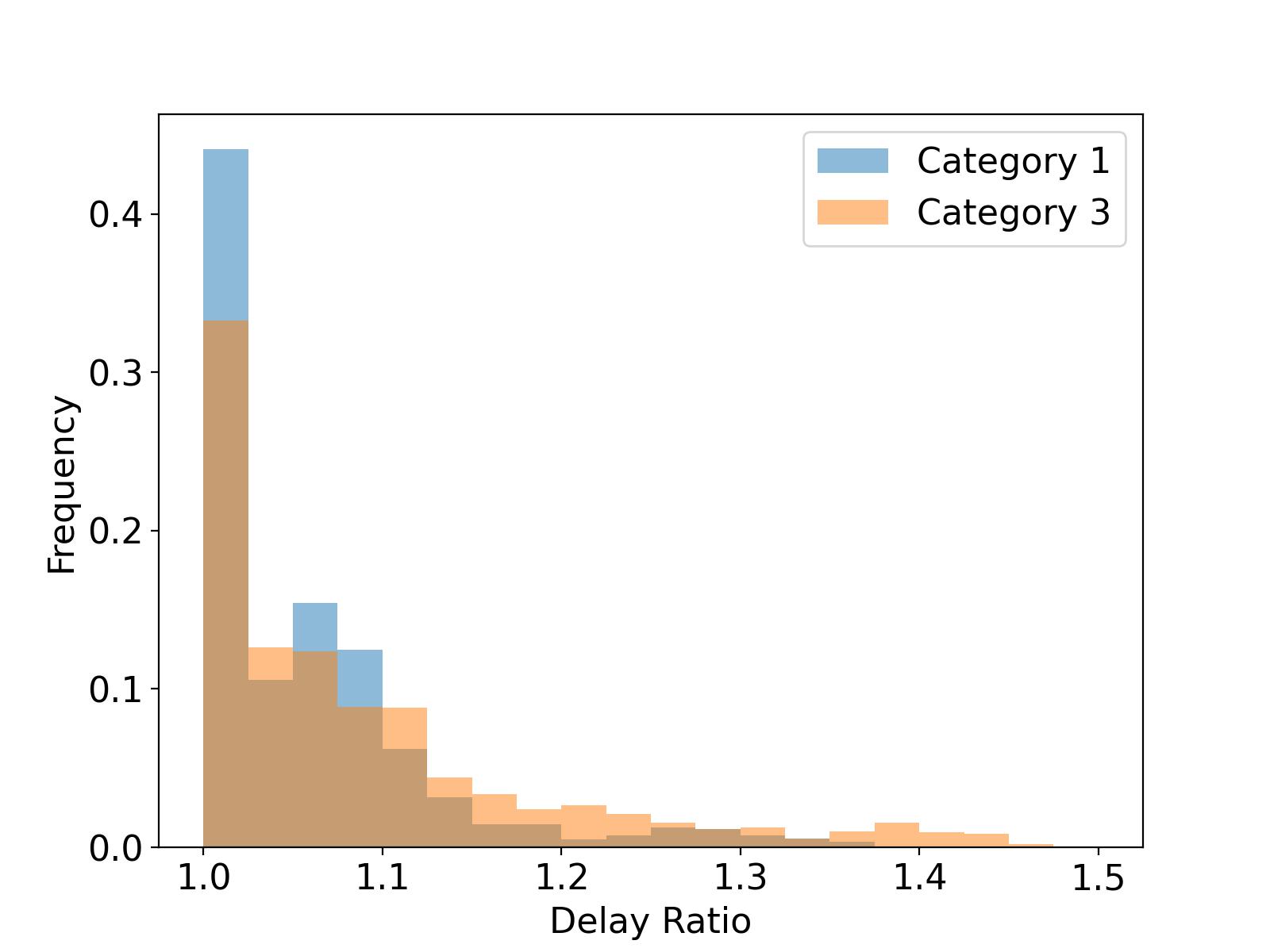}
         \caption{Hillsborough County}
         \label{fig:hist_hillsborough_1_3}
     \end{subfigure}
        \caption{Histogram of average delay ratios distribution at different areas under flood Category 1 and 3.}
        \label{fig:hist_1_3}
\end{figure}

During floods, because of the restricted mobility, and limited visibility, we assumed the traveling speeds of citizens and emergency response vehicles in the flooded area are reduced to 1/3 of their normal values. This leads to two separate roadway network models, for before and after the flood, where edge features are different. For each region, the delay ratio in a selected location, $\Delta$, is calculated by averaging the delay ratios with respect to all the public shelters:

\begin{equation}
    \Delta = 1/K\sum_{i=1}^{K} \delta_i.
\end{equation}

The average delay ratios at different locations under  Category 1 and 3 hazards are shown on the map, as shown in Figure~\ref{fig:time_1} and~\ref{fig:time_3}, respectively. The lighter color indicates a higher average delay ratio. The average delay ratio of the Manhattan area increases from 1.02 to 1.08 from Category 1 to Category 3. For Hillsborough County, the average delay ratio increases from 1.06 to 1.09. The delay ratio distribution can also be shown in Figure~\ref{fig:hist_1_3} under different categories. The heavy tail in the histogram indicates that Category 3 hazards impact a greater number of regions. The most affected areas are found in coastal regions, which align with the flood map. The delay ratio under Category 3 in the most impacted areas of Manhattan and Hillsborough County are 1.30 and 1.48, respectively.  

\begin{figure}[htb!]
     \centering
     \begin{subfigure}[htb!]{0.4\textwidth}
         \centering
         \includegraphics[width=\textwidth]{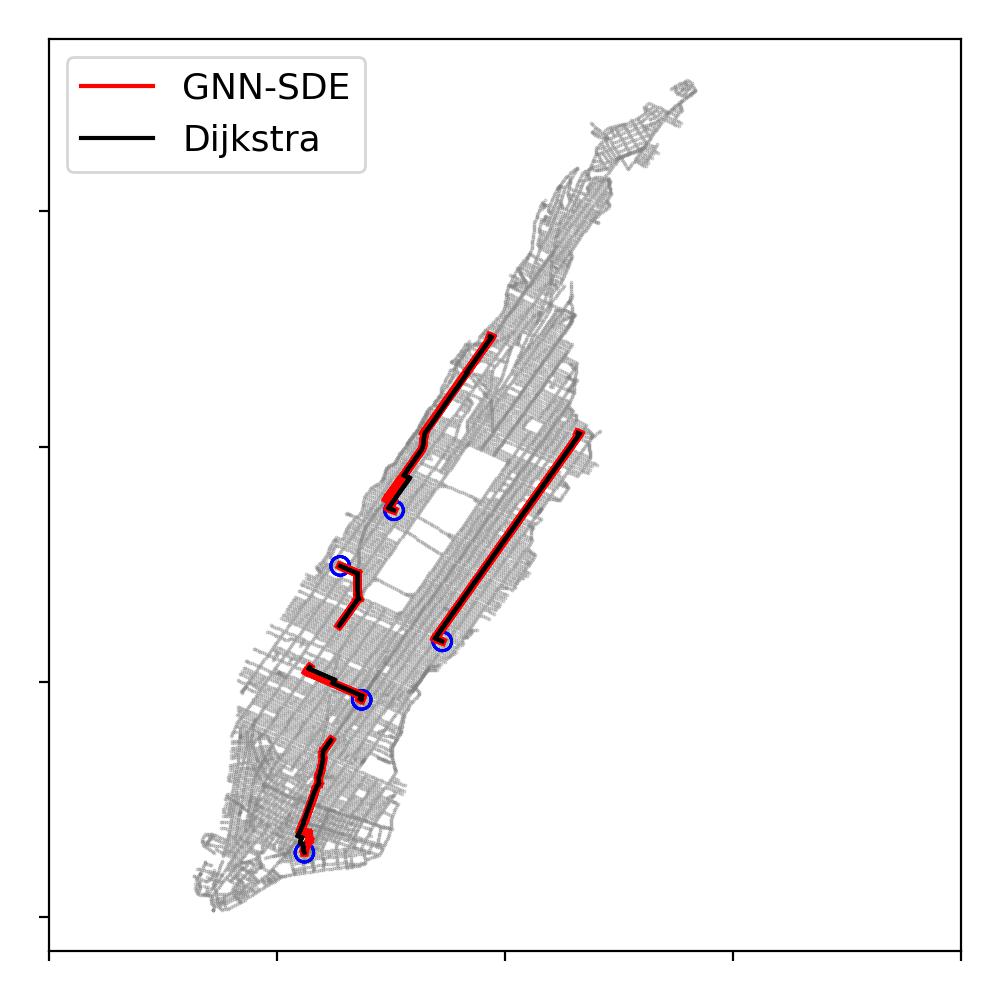}
         \caption{Manhattan}
         \label{fig:manhattan_flood_path}
     \end{subfigure}
     \begin{subfigure}[htb!]{0.4\textwidth}
         \centering
         \includegraphics[width=\textwidth]{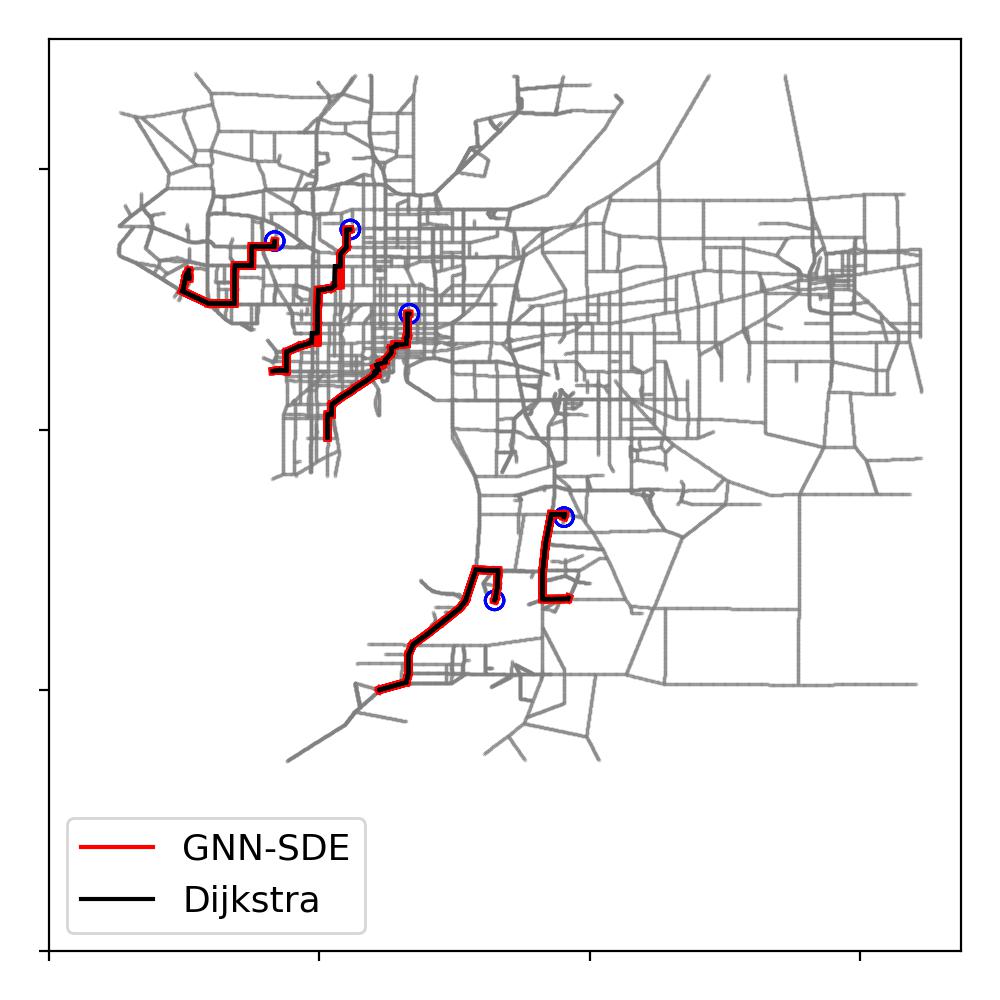}
         \caption{Hillsborough County}
         \label{fig:hillsborough_flood_path}
     \end{subfigure}
        \caption{Shortest path finding and route recommendation comparison between GNN-SDE and Dijkstra's Algorithm. The shortest path is routed from vulnerable location to nearest shelter under flood category 3.}
        \label{fig:path_finding}
\end{figure}

Furthermore, for each potentially flooded and vulnerable area, it is essential to provide recommended evacuation routes to public shelters. Utilizing shortest distance estimation and the proposed Algorithm \ref{alg:path_finding}, we calculate the shortest paths from all public shelter locations to every node in the affected region. Compared with Dijkstra's Algorithm, the MAPE of the shortest time estimation generated by GNN-SDE is 2.11\% and 1.38\% for Manhattan and Hillsborough County, respectively. Figure \ref{fig:path_finding} illustrates five randomly selected locations with their respective routes to the nearest shelters, showcasing the approach's effectiveness in guiding evacuation planning. The figure highlights that the recommended paths closely align with the shortest path results obtained from Dijkstra's Algorithm, underscoring the feasibility and strong performance of the proposed GNN model in real-world scenarios.

Based on the results obtained, several tasks can be identified as potential next steps, including optimizing shelter locations, planning evacuation strategies, and allocating emergency resources. This case study demonstrates the implications and potential applications of the proposed GNN model in the field of urban planning related to disaster preparedness and response.

\section{Discussion and Conclusion}
\label{sec:conclusion}
Efficient estimation of the shortest distance is a fundamental problem with applications in transportation management, emergency services, and network optimization. Traditional approaches, while accurate, often face scalability challenges when applied to large-scale graphs due to their high computational costs. To address these limitations, this paper presents a GNN-based method for shortest distance estimation. The proposed model leverages the representational power of GNNs to capture the spatial relationships and structural information within graphs, achieving both high accuracy and computational efficiency. Compared to conventional methods such as landmark-based techniques and node2vec embeddings, the GNN model demonstrates superior performance in terms of estimation accuracy, while requiring relatively short training times. This combination of efficiency and accuracy positions the GNN approach as a practical alternative for large-scale graph applications.

Numerical experiments conducted on synthetic datasets highlight the model's generalization capabilities with varying topologies. Moreover, its performance on real-world coastal networks illustrates its potential for critical use cases, such as managing transportation during natural disasters. Specifically, the model's ability to estimate the shortest distances between vulnerable areas and public shelters can significantly aid in optimizing evacuation plans, reducing casualties, and enhancing public safety. 

Future work could also explore the application of the framework to multi-modal transportation networks, where the integration of different modes of transport requires advanced modeling of interdependencies and transfer points. Such extensions would further enhance the framework's versatility and applicability in diverse transportation and network optimization contexts. By addressing these directions, the proposed GNN-based framework could not only improve performance in shortest distance estimation but also provide robust, scalable solutions for broader challenges in transportation and emergency response planning. Additionally, the incorporation of uncertainty quantification into the model, using techniques such as Gaussian processes or Bayesian networks \citep{xu2023agnp,zhu2023bayesian}, could provide valuable insights into the reliability of travel time estimates. This would be particularly useful in scenarios with inherent variability, such as during adverse weather conditions or infrastructure failures, where travel times across individual links are subject to random fluctuations.

\section{Acknowledgements}
This work was supported in part by the National Science Foundation under Grant CMMI-1752302.

\bibliographystyle{elsarticle-harv} 
\bibliography{cas-refs}

\end{document}